%% file: acl_latex.tex
\pdfoutput=1

\documentclass[11pt]{article}

\usepackage[final]{acl}
\usepackage{times}
\usepackage{latexsym}

\usepackage{hyperref}

\usepackage[T1]{fontenc}

\usepackage[utf8]{inputenc}

\usepackage{microtype}

\usepackage{inconsolata}

\usepackage{graphicx}

\usepackage{times}
\usepackage{latexsym}
\usepackage{url}
\usepackage{array}
\usepackage{caption}
\usepackage[font=small,skip=0pt]{caption}
\usepackage{afterpage}
\usepackage{inconsolata}
\usepackage{subcaption}
\usepackage{adjustbox}
\usepackage{graphicx}
\usepackage{arydshln} 
\title{Implicit Discourse Relation Classification For Nigerian Pidgin
}

\author{Muhammed Saeed \and Peter Bourgonje \and Vera Demberg \\
  Saarland University, Saarbrücken, Germany \\
  \texttt{\{musaeed,peterb\}@lst.uni-saarland.de, vera@coli.uni-saarland.de, }
  }

\begin{document}
\maketitle

\input{sections/abstract}
\input{sections/introduction}

\input{sections/related}

\input{sections/data}

\input{sections/models}
\input{sections/results}

\input{sections/conclusion}

\input{sections/limitations}


\bibliography{anthology,custom}
\bibliographystyle{acl_natbib}


\appendix
\input{sections/appendix/appendix_a}
\input{sections/appendix/appendix_b}

\section{Appendix E: Full results table}
\input{tables/ablation_table}

\label{sec:appendix}

\end{document}

%% file: sections/abstract.tex
\begin{abstract}


Despite attempts to make Large Language Models multi-lingual, many of the world's languages are still severely under-resourced. This widens the performance gap between NLP and AI applications aimed at well-financed, and those aimed at less-resourced languages. In this paper, we focus on Nigerian Pidgin (NP), which is spoken by nearly 100 million people, but has comparatively very few NLP resources and corpora. 
We address the task of Implicit Discourse Relation Classification (IDRC) and systematically compare an approach translating NP data to English and then using a well-resourced IDRC tool and back-projecting the labels versus creating a synthetic discourse corpus for NP, in which we translate PDTB and project PDTB labels, and then train an NP IDR classifier. The latter approach of learning a "native" NP classifier outperforms our baseline by 13.27\% and 33.98\% in f$_{1}$ score for 4-way and 11-way classification, respectively.

\end{abstract}

%% file: sections/introduction.tex
\section{Introduction}
\label{sec:intro}

An important aspect of understanding a text is correctly parsing the relations between the sentences that compose it, also known as \textit{discourse relations}. 
Uncovering these relations (a task referred to as \textit{discourse parsing}) helps with down-stream tasks such as argument mining \cite{KirschnerEtAl:15}, summarization \cite{xu-etal-2020-discourse,dong-etal-2021-discourse} and relation extraction \cite{tang-etal-2021-discourse}. Recent experiments, however, have shown that discourse parsing performance is not easily improved by modern prompting methods \cite{chan-etal-2024-exploring,yung-etal-2024-prompting}. In addition, multi-lingual LLMs still have comparatively low performance on low-resource languages \cite{zhu2023multilingual,bang-etal-2023-multitask}.

In this paper, we aim to address this by working on discourse parsing for Nigerian Pidgin (NP), which has nearly 100 million speakers, but despite this, has very little support in terms of NLP resources and corpora. We zoom in on implicit discourse relations (following the Penn Discourse TreeBank (PDTB) paradigm \cite{Prasad08thepenn, webber2019penn}, see also Section \ref{sec:data}), and experiment with different methods to classify implicit discourse relations in NP. Particularly, we explore two main strategies: 
\begin{itemize}
    \item The first strategy is based on zero-shot learning. By using a state-of-the-art classifier trained on English \cite{chan-etal-2023-discoprompt}, we both apply the classifier to NP sentences directly, and translate NP text (using machine translation) to English, use the English classifier, and project the annotations back onto the original, NP text. This procedure has the advantage that no annotations in NP are required for training.
    
    \item The second strategy is based on fine-tuning the LLM that is used by the model for relation classification on NP specifically, such that we end up with a dedicated NP model. For this, annotated training data in the target language is required, but we find that this yields better results in the case of NP implicit discourse relation classification. To obtain the annotations for fine-tuning a dedicated NP model, we experiment with two different strategies: 
    \begin{itemize}
        \item By translating the entire text and obtaining the relations and their arguments through word alignment, we preserve context, but risk losing (parts of) relations due to word alignment errors. 
        \item By translating the arguments of individual relations in isolation, alignment becomes trivial and does not result in loss of relations, but by missing the context, translation quality might be lower.
    \end{itemize}
    For translation, we rely on a state-of-the-art English-NP translation system \cite{lin2023low}. For word alignment, we use SimAlign \cite{sabet2020simalign} and AWESoME \cite{dou2021word}, and attempt to improve performance for both using specialized tuning strategies. 
\end{itemize}

We evaluate our approach on gold annotations \textcolor{black}{\cite{scholman2024disconaija}}, and our best-performing set-up achieves an accuracy and f$_{1}$ score of (0.631, 0.461) and (0.440, 0.327), respectively, on 4-way and 11-way relation sense classification (see Section \ref{sec:data} for details).

The synthetic NP corpora with implicit discourse relations used to fine-tune the LLM on which the discourse relation classifier is based, together with the code to run our experiments, is published on GitHub. By disclosing the number of instances we use for fine-tuning an LLM for NP specifically and the relation distribution in both our synthetic (training) and gold (evaluation) NP data, we aim to provide an idea of what performance can be expected for implicit discourse relation classification on a low-resource language such as NP. 

%% file: sections/related.tex
\section{Related Work}
\label{sct:related}

The next sections provide an overview of related work on annotation projection and Implicit Discourse Relation Classification (IDRC). 

\subsection{Annotation Projection}

\paragraph{Discourse Relation Projection} Prior work projecting discourse relation annotations for French, Czech and German is presented by \citet{laali2017automatic, czechpdtb, germanpdtb}. \citet{laali2017automatic} focus on discourse connectives occurring in EuroParl \cite{koehn2005europarl} texts and map their relation senses, taken from a French connective lexicon \cite{lexconn}, to PDTB senses. \citet{czechpdtb} created the Czech PDTB using the Prague Czech–English Dependency Treebank and used human translations of the PDTB3 \cite{pdtb3.0} in combination with Giza++. \citet{germanpdtb} use a similar procedure on German, except that unlike \citet{czechpdtb}, they use automatically obtained translations from DeepL. Of these three, only \citet{germanpdtb} train a classifier on the synthetic data, obtained through annotation projection. In our work, we compare a classifier trained on synthetic data to a zero-shot set-up, the latter being made feasible through improved multi-lingual capabilities of state-of-the-art language models.

With respect to Nigerian Pidgin, a notable contribution is \citet{marchal2021semi}, who focus on creating a lexicon of NP connectives, by exploiting a parallel corpus \cite{caron2019surface} and discourse parsing \cite{LinEtAl:14}. In contrast to \citet{marchal2021semi}, who focus on explicit markers of discourse relations (\textit{discourse connectives}), we focus on implicit discourse relations.

\paragraph{Machine Translation for NP}
Generally, annotation projection is used in a scenario where one language has considerably more resources or better task performance than the language of interest. By translating input in this language to the better-resourced language and running tools on the translated text, the annotations can be projected back onto the original text by leveraging the character offsets obtained through word alignment.
Annotation projection thus typically starts with (machine) translation.
Particularly relevant for our work is the AfriBERTa translation model proposed by \citet{ogueji2021small}, which includes English-NP as a language pair. \citet{pcm_paper_baseline} specifically target NP, and presented the first neural translation system for NP using a Transformer model \cite{vaswani2017attention} trained on 27k sentences from JW300 \cite{agic-vulic-2019-jw300}, without prior transfer learning. \citet{lin2023low} enhanced this approach by including bible data sets and applying transfer learning with the T5 model \cite{raffel2020exploring}. \citet{lin2023low} found that English versions of T5 and RoBERTa outperformed their multi-lingual equivalents on NP, presumably due to English being the lexifier for NP.
\citet{tan2021msp} focus on a more resource-efficient approach dubbed Multi-Stage Prompting, demonstrating its efficiency on Romanian-English, English-German and English-Chinese translations. In this paper, we combine \citet{tan2021msp} and \citet{lin2023low} to translate the implicit relations of the original, English PDTB into NP, as a (synthetic) training corpus for IDRC in NP. 

\paragraph{Word Alignment} Once two versions (in different languages) of the same text exist, a word alignment procedure can be used to link these version at word- and character-level, allowing for annotations to be projected back and forth. In recent years, traditional methods employing HMM models \cite{schonemann1966generalized,brown1993mathematics,och2000improved,och2003systematic} have been superseded by neural methods. We include a Python version of the original GIZA++ aligner \cite{10.1162/089120103321337421}, but we focus more on neural methods in our work. Specifically, \citet{sabet2020simalign} propose SimAlign, using multi-lingual embeddings. Their methods —Argmax, IterMax, and Match— offer different recall and precision balances, with sub-word processing proving beneficial for aligning rare words. \citet{dou2021word} introduce AWESoME (Aligning Word Embedding Spaces of Multilingual Encoders), an architecture that combines pre-trained language models such as BERT and RoBERTa with finetuning. We experiment with SimAlign and AWESoME and attempt to improve performance by fine-tuning the underlying models for NP.

\subsection{Implicit Discourse Relation Classification}
This paper aims to contribute to IDRC for low-resource languages, using NP as a case study. 
We use DiscoPrompt \cite{chan-etal-2023-discoprompt} throughout our work (as a baseline, but also as a basis for further training). DiscoPrompt (``Discourse relation path prediction Prompt tuning model'') incorporates the hierarchy of the PDTB into prompts to its pre-trained model (T5 \cite{raffel2020exploring}), to jointly predict both top-level and second-level relation senses (and connectives as well, but we do not use this information in our paper). We adopt DiscoPrompt for its state-of-the-art results, robustness and ease of use.

Like DiscoPrompt, a considerable amount of other prior work on IDRC in recent years \cite{shi-demberg-2019-next,kishimoto-etal-2020-adapting,liu2020importance,wu2022label} 
has focused on English.
An attempt to include the multi-lingual perspective, however, has been put forward by the Discourse Relation Parsing and Treebanking (DISRPT) shared task series \cite{zeldes-etal-2019-disrpt,zeldes-etal-2021-disrpt,braud-etal-2023-disrpt}. In this context, some other approaches have been suggested, such as DiscoFlan \citet{anuranjana-2023-discoflan}, which transforms IDRC into a label generation task using the FlanT5 model, and uses instruction fine-tuning in multi-lingual settings. \citet{metheniti-etal-2024-zero-shot} assess multi-lingual BERT's cross-lingual transfer learning capabilities across different languages and frameworks (PDTB \cite{Prasad08thepenn} and RST \cite{MannThompson:88}). \citet{liu-etal-2023-hits} train individual classifiers (comprising pre-trained models as encoders and linear networks as classification layers) for larger corpora, but employ a joint model for smaller datasets. \citet{gessler-etal-2021-discodisco} present DiscoDisco, which utilizes a feature-rich, encoder-less sentence pair classifier for relation classification.
Outside of shared tasks, \citet{bourgonje-lin-2024-projecting} deploy a multi-lingual discourse parsing pipeline, evaluating it on 5 languages, focusing solely on discourse connectives.

\citet{kurfali-ostling-2019-zero} contributed to multi-lingual IDRC (and less-resourced languages), through zero-shot learning using language-agnostic models like LASER \citet{artetxe-schwenk-2019-massively}, to classify discourse relations in Turkish, which has relatively little training data. All multi-lingual approaches mentioned above, however, use a zero-shot transfer learning set-up. By contrast, \citet{germanpdtb} train on synthetic (what they refer to as ``silver'') data in German. To the best of our knowledge, we are the first to compare both a zero-shot transfer set-up to an approach based on training a classifier with synthetic data for the IDRC task, and the first to work on IDRC for NP.

%% file: sections/data.tex
\section{Data}
\label{sec:data}

\begin{figure}[htbp]
    \centering
    \includegraphics[width=\columnwidth]{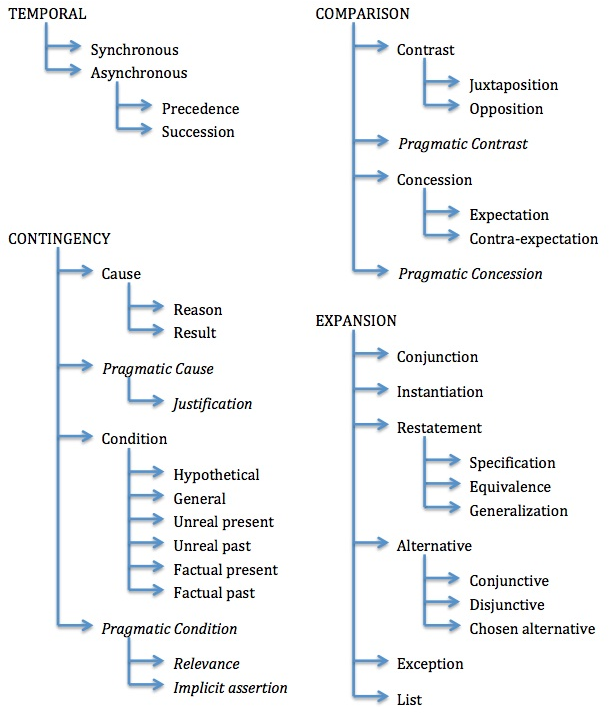}
    \caption{PDTB2.0 Sense Hierarchy}
    \label{fig:pdtb2_inventory}
\end{figure}

Our final goal is to classify \textit{implicit} discourse relations in text. For this, we follow the PDTB framework \cite{Prasad08thepenn,pdtb3.0}. In the PDTB, which is annotated over financial (Wall Street Journal) news articles, discourse relations are first categorized into one of several types, of which \textit{explicit} and \textit{implicit} relations are the most frequent. In the former, relations are explicitly and lexically signalled by a discourse connective, words and phrases such as ``however'', ``as a result'' and ``either ... or''. In the latter, such explicit signals are lacking, and the reader must rely on the semantics of the two arguments of the relation to infer the relation \textit{sense}. Every relation instance has exactly two arguments (``Abstract Objects'' \cite{Asher:93}, referred to as \textit{arg1} and \textit{arg2}), and a relation sense. In our evaluation set-up, our classifier takes these two arguments, and outputs a particular relation sense. For the sense inventory, we adopt the PDTB2 version of the sense hierarchy, which is illustrated in Figure \ref{fig:pdtb2_inventory}. 
We follow related work in classifying the top-level senses (4-way classification) and second-level senses (theoretically, 16-way classification, but since not all sense occur in the PDTB data, practically 11-way classification). 

\subsection{Evaluation Data}
\label{ssec:eval_data}
\begin{figure}[h]
    \centering
    \includegraphics[height=0.2\textheight, width=\columnwidth]{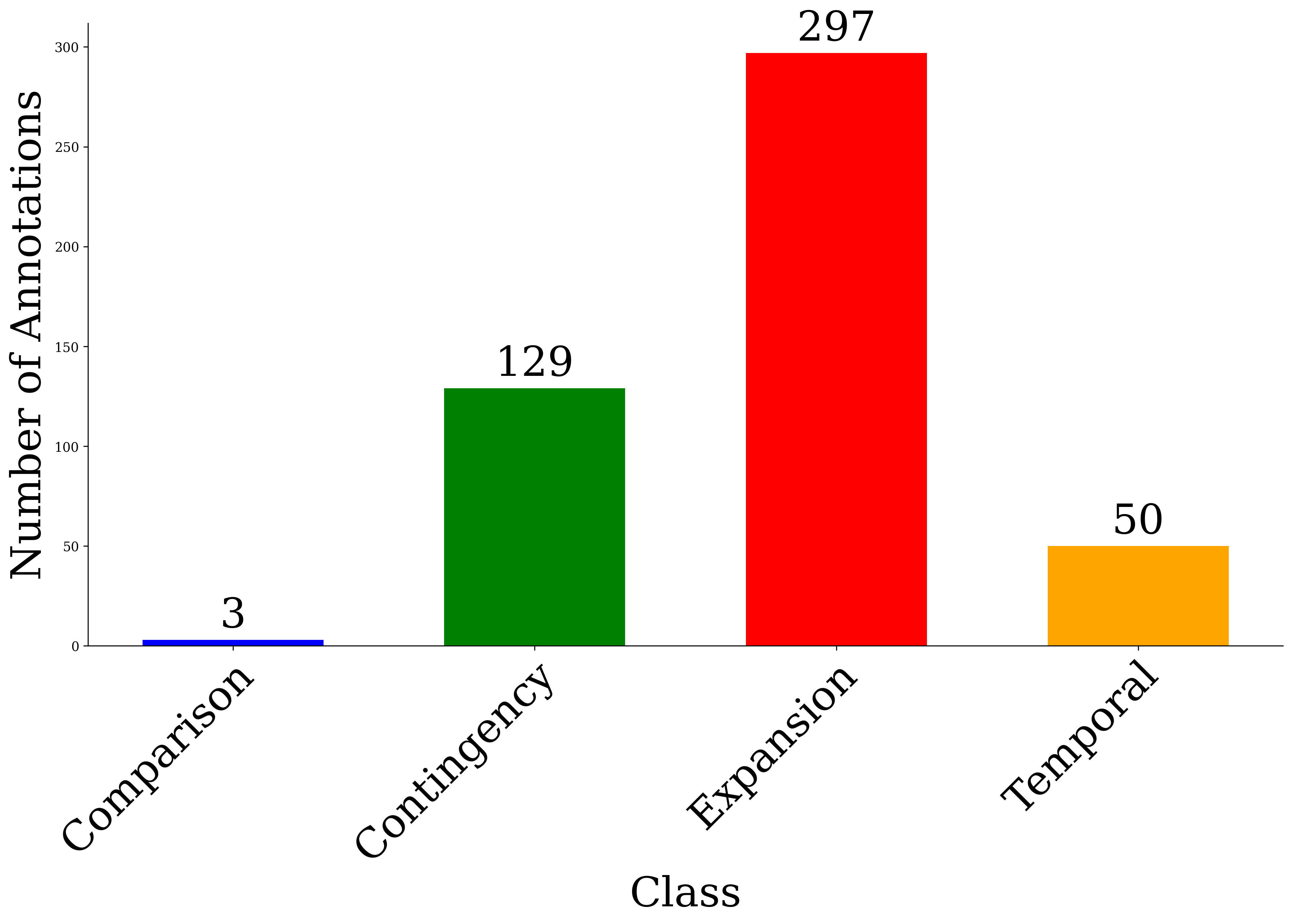}
    \includegraphics[height=0.2\textheight, width=\columnwidth]{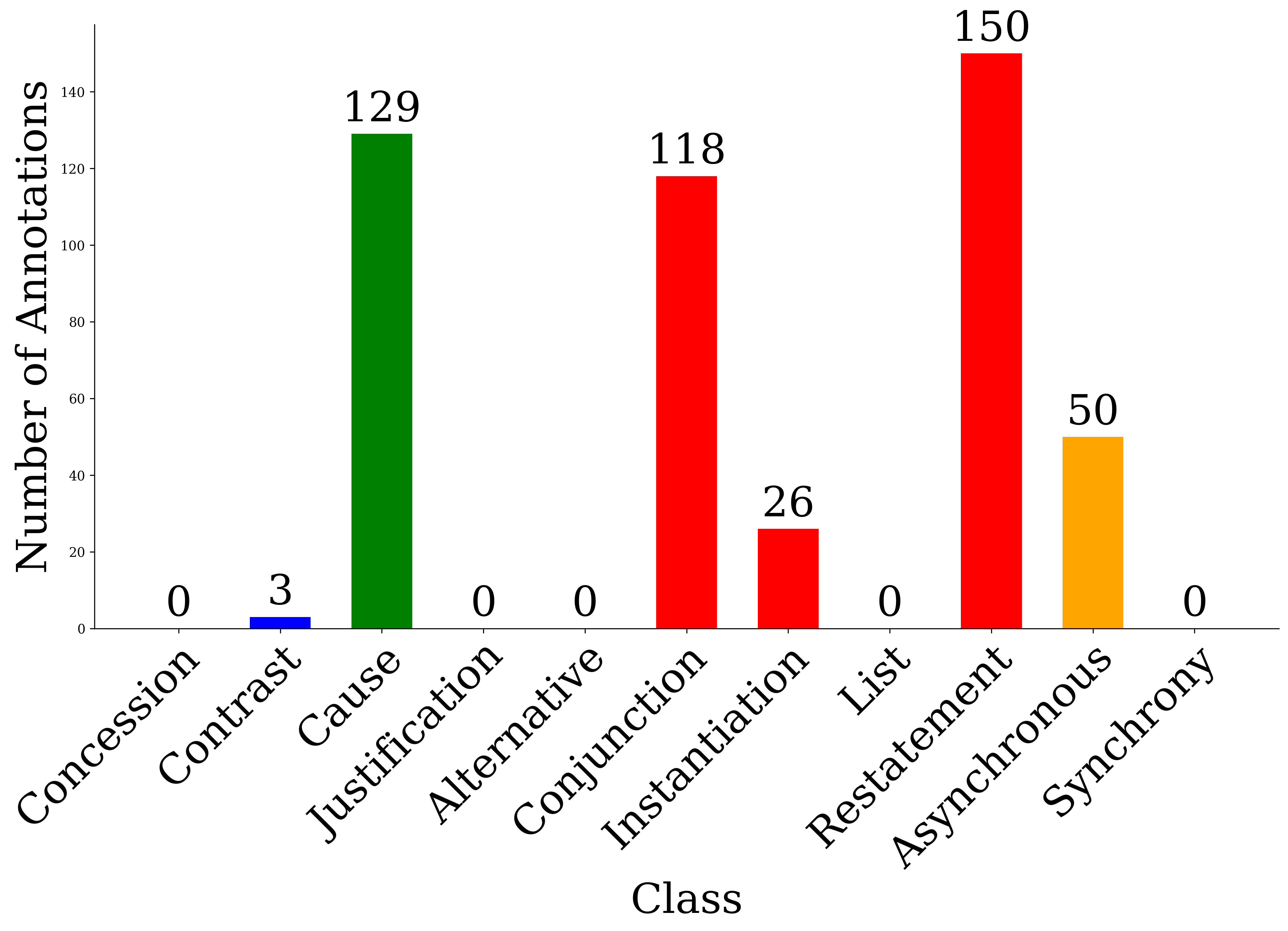}
    \caption{Test-set Relation Sense Distribution.}
    \label{fig:testset SenseDistributions}
\end{figure}

For both our zero-shot learning based strategy and our LLM fine-tuning based strategy, we use the corpus provided by \citet{scholman2024disconaija}\footnote{Manuscript in preparation, data set obtained through personal communication.}, which is annotated according to PDTB guidelines. This is based on the data set provided by \citet{caron2019surface}, which features 500k words from 321 audio files collected across 11 locations in Nigeria, encompassing diverse discourse types like speech and radio programs. \citet{scholman2024disconaija} annotated a sub-set of this, comprising approximately 140k words, with 12,274 relation tokens. We use the test set of \citet{scholman2024disconaija}, which contains 601 implicit relations. The distribution of relation senses in the test set on both levels of the PDTB sense hierarchy are illustrated in Figure \ref{fig:testset SenseDistributions}.

Since our English data consists of (financial) news, and our evaluation data contains radio program discussions and life narratives, we are dealing with a domain transfer evaluation set-up.

\subsection{Unannotated and Synthetic Data} 
\label{ssec:syn_data}

In order to fine-tune the LLM that our relation classifier is based on, we need NP implicit relation annotations. This is a two-step process, for which we first need unannotated, plain NP text, to continue training the embeddings model that is used by the word alignment method. After the word alignment method has been enhanced, we generate a synthetic, NP version of the original PDTB to train our relation classifier.

\paragraph{Plain NP text}
To fine-tune the embedding model used by the neural word aligners, AWESoME \cite{dou2021word} and SimAlign \cite{sabet2020simalign}, we need parallel sentences. For AWESoME, we use a method called parallel fine-tuning (PFT), in which we take a pre-trained mBERT model fine-tune this on the approx. 48k parallel English-NP sentences from \citet{lin2023low}, which in turn are coming from \citet{pcm_paper_baseline} and represent texts from the religious domain. As SimAlign does not support the process of fine-tuning with parallel sentences, we modify SimAlign by integrating the RoBERTa model from \citet{lin2023low}, which was fine-tuned using a method called cross-lingual adaptive training (CAT) with approximately 300K monolingual NP sentences.

\begin{figure}[h!]
    \centering
    

        \includegraphics[width=\columnwidth,  trim=0cm 15cm 0cm 1cm, clip=true]{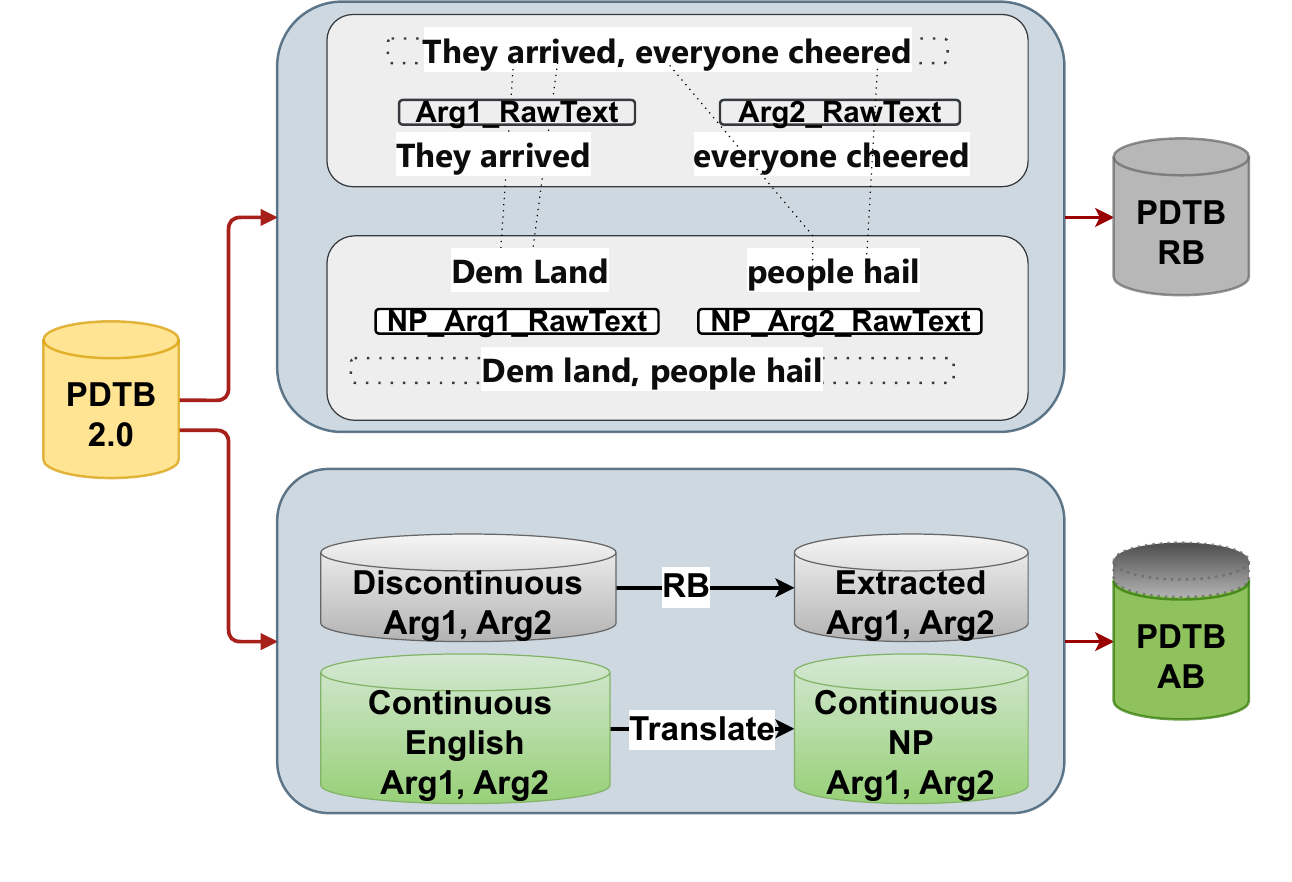}  
        \caption{In the fine-tuning approach, the upper part illustrates the relation-based (RB) method with parallel English and NP sentences to align and extract arguments in NP. The lower part illustrates the argument-based (AB) method, directly translating continuous arguments and using alignment (essentially, the RB method) for (a relatively small number of) discontinuous ones. 
        The AB method relies less on alignment, resulting in fewer lost arguments/relations, hence the NP PDTB AB dataset is larger than the NP PDTB RB dataset. }
        \label{fig:second_figure}
    
\end{figure}

\paragraph{NP PDTB}
\label{sec:NP PDTB}
To arrive at an NP PDTB, we try two different strategies (illustrated in Figure \ref{fig:second_figure}):

\paragraph{Relation-based translation (RB)}
First, we translate the original English text of the entire relation into NP and then apply word alignment to find the two arguments. The translation is done with the model provided by \citet{lin2023low}. We then rely on the output of different word alignment tools we use to combine character offsets from the original English annotations with word (and ultimately character) alignments from the aligner tool to end up with NP relations.
This way, more context is preserved because we translate the entire relation at once. However, errors during word alignment may result in certain arguments not being found in the NP text. We call this method relation-based translation.

\paragraph{Argument-based translation (AB)}
Second, we translate individual arguments of relations (\textit{arg1} and \textit{arg2}) in isolation. We also use the model from \citet{lin2023low} for this, but feed it individual arguments
, rendering alignment trivial in most cases: The output, in most cases, is the NP relation, consisting of two arguments and the sense directly taken from the English seed relation. The cases in which alignment is not trivial, concern discontinuous argument spans. Upon analysis, we found that of the 16,053 implicit relations in the PDTB, only 1,180 have discontinuous argument spans. 
For discontinuous arguments, we first translate the entire token span from the first to the last word of the argument, and then exploit word alignment output to exclude words that are not part of the argument.

In addition to these to different strategies, we also compare three different alignment tools. These are Giza-py, which is a pre-neural alignment model, as well as two neural
word aligners (AWESoME and SimAlign). 
These word alignment tools are of course also not trained on NP, therefore, we experiment with the original English tools, as well as with NP fine-tuned versions: AWESoME+PFT and SimAlign+CAT. The AWESoME aligner allows fine-tuning on parallel datasets, termed "AWESoME+PFT", we finetune the model using 48K \cite{lin2023low, agic-vulic-2019-jw300}. SimAlign aligner is based on mBERT, yet according to \cite{lin2023low}, they have fine-tuned English RoBERTa \cite{liu2019roberta} model on 300K monolingual NP sentences and achieved state-of-art results in sentiment analysis, we have used \cite{lin2023low} engine with SimAlign to generate Alignment and called it SimAlign+CAT.

The original PDTB contains 16,053 instances of implicit relations. The resulting corpus for the argument-based Translation strategy has almost the exact same number of relations. However, due to errors in the alignment process, not all instances are successfully mapped to the NP text, resulting in some relations getting lost. 
Table \ref{tab: full transfer data-stats} provides an overview of the number of relations per method (argument-based Translation Translation and argument-based Translation) and per corpus (Giza-py, SimAlign (with and without tuning) and AWESoME (with and without tuning)). As illustrated, the argument-based Translation method preserves all implicit relations in the case of Giza-py and AWESoME, but loses some (discontinuous) instances in the case of SimAlign. The argument-based Translation Translation method loses significantly more relation instances, but presumably, context is better preserved in the translations.
Further details, including the distribution of relation senses, are included in the Appendix. 

\begin{table}[!ht]
\centering
\begin{tabular}{l|c|c}
\hline
Corpus      & AB \# rel. & RB \# rel. \\ \hline
Giza-py     & 13,022 & 16,053       \\ 
SimAlign    & 13,201 & 15,527      \\ 
SimAlign+CAT & 13,206 & 15,531       \\
AWESoME      & 12,975 & 16,053      \\ 
AWESoME+PFT  & 12,943 & 16,053     \\ 
\hline
\end{tabular}
\caption{Number of implicit relations in different versions of our NP PDTB corpus. AB = argument-based Translation Translation, RB = argument-based Translation.}
\label{tab: full transfer data-stats}
\end{table}

%% file: sections/models.tex
\section{Set-ups}
\label{sec:models}

\begin{figure}[h!]
    \centering
    
        \includegraphics[width=\columnwidth,   trim=0cm 20cm 0cm 0cm, clip=true]{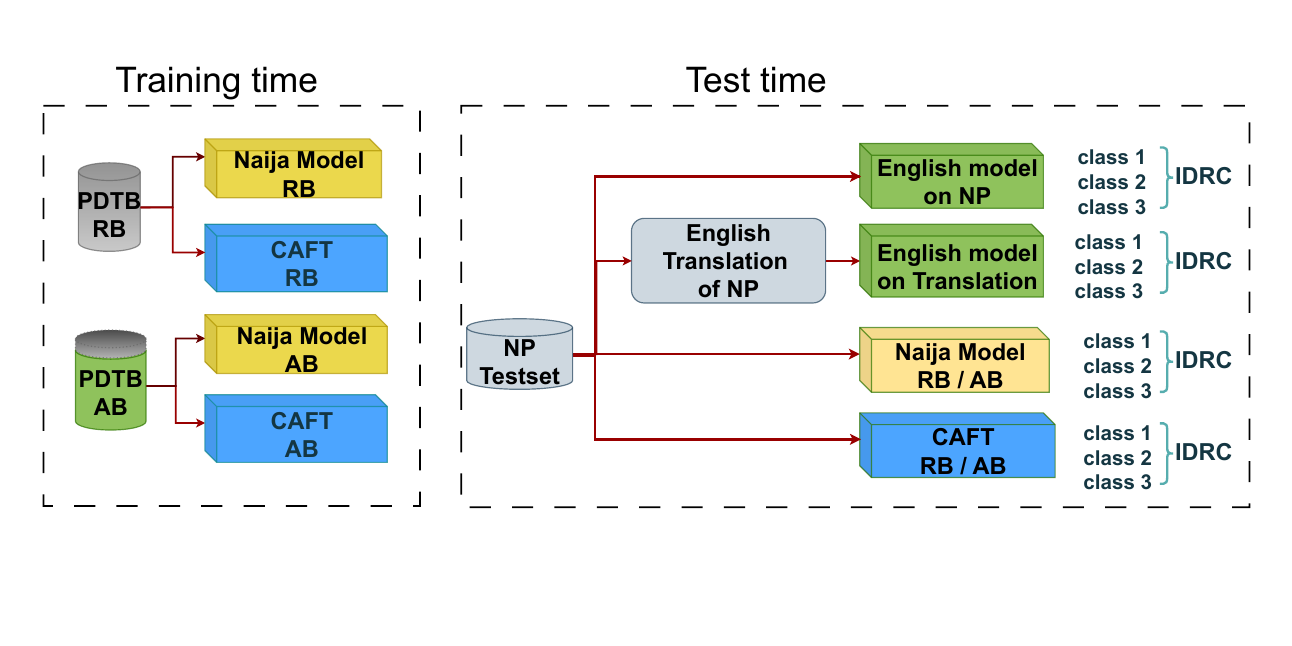}
        \caption{Illustration of the four different approaches outlined in Sections \ref{sec:EnglishIDRConNP} and \ref{sec:FinetuningusingNPPDTB}.
        }
        \label{fig:framework}
\end{figure}

With our final goal being Implicit Discourse Relation Classification (IDRC) for Nigerian Pidgin (NP), this section outlines the four experimental set-ups employed for this task, which all utilize the state-of-the-art DiscoPrompt model \cite{chan-etal-2023-discoprompt}. We evaluate both the base and large variants of DiscoPrompt across all set-ups using the gold data described in Section \ref{ssec:eval_data}. The set-ups include two methods which use the English DiscoPrompt model and two approaches that train / fine-tune on the NP PDTB dataset; they are illustrated in Figure \ref{fig:framework}. We obtained the best results with a version of the NP PDTB resulting from our relation-based method, and thus use this version for the two methods that leverage fine-tuning on NP annotations. We present more detailed results and discuss the difference of both NP PDTB corpus creation strategies in Section \ref{sec:results}.

\subsection{English IDRC on NP}
\label{sec:EnglishIDRConNP}
The first category we have used for the IDRC is English IDRC on NP, which is Zero-shot from English Model trained  IDRC on the NP dataset and we are employing two scenarios the first is directly on the NP and the second on the translation.
\subsubsection{English model on NP}
We use the original, English DiscoPrompt model from \citet{chan-etal-2023-discoprompt} without any modifications as a baseline. Since NP is English-lexified, and earlier work \cite{lin2023low} has shown that for NP, an English model works better than a multi-lingual one, we assume this to be a reasonable baseline.


\subsubsection{English Model on Translation}
In this set-up, we translate the relations in the test data into English, apply the original DiscoPrompt model on the translated arguments, and then evaluate against the gold annotations. In addition to serving as another approach to compare to, this set-up particularly assesses the model's adaptability to translated texts.

\subsection{Fine-tuning using NP PDTB}
\label{sec:FinetuningusingNPPDTB}
These set-ups utilize the relation-based NP PDTB data set described in Section \ref{ssec:syn_data}, by training an NP version of DiscoPrompt. 
We experiment with two different approaches.

\subsubsection{Naija Model (NM) }
\label{subsec:naijamodel}
This approach trains an NP DiscoPrompt model from scratch, using our synthetic NP PDTB.

\subsubsection{Continuous Adaptive Fine-Tuning (CAFT)} 
This approach starts with the original English DiscoPrompt model (which is trained on the English PDTB) and further fine-tunes it on the NP PDTB. We evaluate both the base and large versions of the models underlying DiscoPrompt to determine the model's susceptibility to overfitting on the NP PDTB.


%% file: sections/results.tex
\section{Results}
\label{sec:results}

Our experiments compare a rich set of model settings, including the IDRC method,  T5 model size (base vs.~large), alignment basis (argument-based vs.~relation-based) and five different alignment models. We will discuss the effect of each of these in turn. 
We report the results of the IDRC task on our evaluation data using 11-way f$_1$ and accuracy, as well as 4-way f$1$ and accuracy.
A full table including all results is provided as part of Table \ref{tab:ablation} in the Appendix. 

\paragraph{IDRC method}
We firstly compare the different approaches of using an English model directly on the NP data vs.~translating the NP data to English and then using the English model vs.~training a NP model from scratch vs.~fine-tuning an English pretrained model to NP data. Our results in Table \ref{tab:4_setups} show that this last approach performs best with respect to f$_1$ on the 4-way and 11-way classification, as well as 11-way accuracy. The model results shown in Table \ref{tab:4_setups} show results for the model setting with argument-based translation and the alignment performed using AWESoME+PFT.

For the two English IDRC methods, the one where we feed NP relation instances directly to an English model performs better. This suggests that the model itself is better able to ``internally'' translate the discourse relations, than a state-of-the-art external MT engine is at preserving the cues that are relevant for labelling implicit discourse relations. 

\input{tables/4_setups}

\paragraph{DiscoPrompt T5 model variant}
We also found that performance was generally higher with the Large T5 model variant for the CAFT method, see the top vs.~bottom halves of Table \ref{tab:4_setups} for the model setting with argument-based translation and AWESoME+PFT alignment. 

Interestingly, we also observed that when training the NP IDRC model from scratch, performance was in many settings better when using the base T5 model as a basis for the DiscoPrompt model, see Table \ref{tab:small-large}.

\begin{table}[!ht]
\centering
\begin{adjustbox}{width=\linewidth}
\begin{tabular}{lcccc}
\hline
\textbf{} & \textbf{4-way}& & \textbf{11-way}& \\
\textbf{} & \textbf{f$_{1}$} & \textbf{Acc} & \textbf{f$_{1}$} & \textbf{Acc} \\
\hline
\textbf{large T5 model} \\ \hdashline
\textbf{AWESoME} & 0.159 & 0.340 & 0.113 & 0.317 \\ 
\textbf{AWESoME+PFT} & 0.364& 0.519 & 0.243 & \textbf{0.400} \\
\textbf{SimAlign} & 0.296& 0.521 & 0.195 & 0.327 \\
\textbf{SimAlign+CAT} & 0.324 & 0.445 & 0.167 & 0.335 \\
\textbf{Giza-py} & 0.327& 0.389 & 0.216 &0.310 \\
\hdashline
\textbf{base T5 model} \\ \hdashline
\textbf{AWESoME} & 0.354 & 0.515 & \textbf{0.278} & 0.392\\
\textbf{AWESoME+PFT} & \textbf{0.357} & 0.538 & 0.246 & 0.340 \\
\textbf{SimAlign} & 0.375 & \textbf{0.610} & 0.230 & 0.360 \\
\textbf{SimAlign+CAT} & 0.229 & 0.377 & 0.183 & 0.321 \\
\textbf{Giza-py} &0.335& 0.464 &0.256 &0.360\\
\hline
&\\
\end{tabular}
\end{adjustbox}
\caption{Performance of the DiscoPrompt model trained from scratch on Naija Pidgin PDTB data.}
\label{tab:small-large}
\end{table}

\paragraph{Argument-based vs.~relation-based translation}
We next compare the effect of different translation projection methods for corpus creation (relation-based and argument-based). 
As mentioned in Section \ref{sec:intro}, we expect the relation-based method to result in slightly noisier data compared to the argument-based method, as words might be missing from arguments due to alignment errors. However, we also expect the translation quality for the relation-based method to be slightly better than for the argument-based method, as more context is taken into account during translation. 
Our results in Table \ref{tab:rel-align} show that the high alignment quality obtained from translating each argument separately outweighs the potential benefit from reflecting a larger context in the translation. The misalignment problems cause the model performance to consistently degrade compared to the model that uses the trivial alignment by translating each argument separately.


\begin{table}[!ht]
\centering
\begin{adjustbox}{width=\linewidth}
\begin{tabular}{lcccc}
\hline
\textbf{} & \textbf{4-way}& & \textbf{11-way}& \\
\textbf{} & \textbf{f$_{1}$} & \textbf{Acc} & \textbf{f$_{1}$} & \textbf{Acc} \\
\hline
\textbf{argument-based} \\ \hdashline
\textbf{AWESoME} & 0.433 & 0.585 & 0.325 & 0.444 \\
\textbf{AWESoME+PFT} & 0.437 & 0.573 & \textbf{0.327} & 0.440 \\
\textbf{SimAlign} & \textbf{0.461} & \textbf{0.631} & 0.246 & 0.429 \\
\textbf{Simalign+CAT} & 0.411 & 0.571 & 0.236 & 0.425 \\
\textbf{Giza-py} & 0.431& 0.566 & 0.289 &\textbf{0.446} \\

\hdashline
\textbf{relation-based} \\ \hdashline
\textbf{AWESoME} & 0.408 & 0.575 & 0.304 & 0.416 \\
\textbf{AWESoME+PFT} & 0.371 & 0.481 & 0.289 & 0.400 \\
\textbf{SimAlign} & 0.362 & 0.498 & 0.259 & 0.396 \\
\textbf{SimAlign+CAT} & 0.358 & 0.506 & 0.259 & 0.404 \\
\textbf{Giza-py} & 0.368 & 0.552&0.245 &0.371 \\
\hline
&\\\end{tabular}
\end{adjustbox}
\caption{Performance of the DiscoPrompt model trained from scratch on Naija Pidgin PDTB data. The table shows results for CAFT T5 large models.}
\label{tab:rel-align}
\end{table}

\input{tables/confusionMatrix/AweSoMEPFT_11Way}

\paragraph{Alignment models} 
Finally, we would like to discuss the effect of our different alignment algorithms. Consider again Table \ref{tab:rel-align}. We find that there is no single alignment method that outperforms the other ones across all the different settings. Overall, the neural alignment methods tend to yield slightly better results than Giza-py. For the AWESoME alignment method, we find that performance is typically improved when the aligner is finetuned on NP (AWESoME+PFT), while for SimAlign, we often observe better results for a setting that does not use fine-tuning on NP data.

In summary, our best method (Naija model with CAFT) surpasses our baseline (English model on NP) by 13.27\% in top-level F$_{1}$ and 33.98\% in second-level F$_{1}$ for the large model, and by 12.71\% in top-level F$_{1}$ and 68.89\% in second-level F$_{1}$ for the base model as shown in Table \ref{tab:ablation}.

\subsection*{Error analysis}

From the confusion matrix in Table \ref{tab:confusionmatrix}, we observe a confusion in the 'Conjunction' class, where the model misclassified 'Conjunction' as 'Cause' 41 times and as 'Restatement' 40 times, while correctly identifying 'Conjunction' only 25 times. For 'Instantiation,' the correct label is applied 8 times but a mis-classification as 'Cause' occurs 10 times and as 'Restatement' 6 times. In the case of 'Restatement,' the model correctly classified the corresponding relations 78 times but misclassified it as 'Cause' 54 times. The confusion between 'Restatement/Instantiation' and 'Cause' is known to be challenging also for human annotators \cite{scholman2017examples}. 

The 'Asynchronous' category shows 18 correct classifications but is incorrectly classified as 'Cause' 20 times. Manual analysis of these cases revealed that roughly half of these are possible secondary interpretations, while the other half does not have a good causal interpretation, possibly due to model misinterpretation of time or context.  
Interestingly, the classes 'Cause' and 'Restatement' exhibit a comparatively high number of false positives. This could be related to the distribution in the training data (see Figure \ref{fig:awesomealign_pft_projected}), where 'Cause', 'Conjunction' and 'Restatement' make up the top three of most frequent classes.
Addressing these imbalances with additional training data (from different classes) could thus improve the model's accuracy. 



%% file: tables/4_setups.tex
\begin{table}[!ht]
\centering

\begin{adjustbox}{width=1\linewidth}
\tabcolsep=3pt
\begin{tabular}{lcccc}
\hline
 & \textbf{4-way}& & \textbf{11-way}& \\
\textbf{DiscoPrompt Model}& \textbf{f$_{1}$} & \textbf{Acc} & \textbf{f$_{1}$} & \textbf{Acc} \\
\hline
\textbf{... with Large T5 model} \\ \hline
{EN model on NP} & 0.407 & 0.597 & 0.244 & 0.391 \\
{EN model on translation} & 0.344 & 0.537 & 0.222 & 0.373 \\
{Naija model from scratch}  & 0.364& 0.519 & 0.243 & 0.400 \\ 
{Naija model with CAFT} &  \textbf{0.437} & 0.573 & \textbf{0.327} & \textbf{0.440}  \\ 
\hline
\textbf{... with Base T5 model} \\ \hline
{EN model on NP} & 0.351& \textbf{0.602} & 0.180 & 0.341 \\
{EN model on translation} & 0.339 & 0.545 & 0.240 & 0.342 \\
{Naija model from scratch} &0.357 & 0.538 & 0.246 & 0.340  \\ 
{Naija model with CAFT} & 0.381 & 0.550 & 0.241 & 0.396 \\ 
\hline
&\\
\end{tabular}
\end{adjustbox}

\caption{Results for the four main set-ups, see model descriptions in Sections \ref{sec:EnglishIDRConNP} and \ref{sec:FinetuningusingNPPDTB}. 
Alignments for the models displayed here are obtained with AWESoME+PFT and argument-based translation. Best scores are marked in bold.}
\label{tab:4_setups}
\end{table}

%% file: tables/confusionMatrix/AweSoMEPFT_11Way.tex
\begin{table*}[h!]
\centering
\begin{adjustbox}{width=\textwidth}
\begin{tabular}{c|c|c|c|c|c|c|c|c|c|c|c}
\textbf{True \textbackslash Prediction} & \textbf{Conc.} & \textbf{Contr.} & \textbf{Cause} & \textbf{Just.} & \textbf{Alt.} & \textbf{Conj.} & \textbf{Inst.} & \textbf{List} & \textbf{Restat.} & \textbf{Async.} & \textbf{Sync.} \\
\hline
\textbf{Concession} & & & & & & & & & & & \\ \hline
\textbf{Contrast} & & \underline{2} & & & & 1 & & & & & \\ \hline
\textbf{Cause} & & 9 & \underline{80} & & 2 & 5 & 2 & & 30 & 3 & \\ \hline
\textbf{Justification} & & & & & & & & & & & \\ \hline
\textbf{Alternative} & & & & & & & & & & & \\ \hline
\textbf{Conjunction} & & 6 & 41 & & & \underline{25} & 1 & & 40 & 4 & \\ \hline
\textbf{Instantiation} & & 1 & 10 & & & 1 & \underline{8} & & 6 & & \\ \hline
\textbf{List} & & & & & & & & & & & \\ \hline
\textbf{Restatement} & & 3 & 54 & & 1 & 9 & 2 & & \underline{78} & 4 & \\ \hline
\textbf{Asynchronous} & & 3 & 20 & & & 6 & & & 5 & \underline{18} & \\ \hline
\textbf{Synchronous} & & & & & & & & & & & \\ 
\end{tabular}
\end{adjustbox}
\caption{11-way classification confusion matrix for the Naija Model DiscoPrompt T5 Large with CAFT on the dataset generated by AWESoME+PFT.}
\label{tab:confusionmatrix}
\end{table*}


%% file: sections/conclusion.tex
\section{Conclusion}
\label{sec:conclusion}

In this paper, we present an approach to Implicit Discourse Relation Classification for Nigerian Pidgin (NP). We experiment with zero-shot transfer, by 1) applying an English classifier model on NP directly, and by 2) translating input text into English and projecting the results back onto the NP source text. We find, however, that a dedicated NP classifier, trained on synthetically generated NP discourse relation annotations, outperforms both zero-shot transfer set-ups. We try both an NP model trained from scratch on our synthetic NP annotations and an English discourse relation classifier that we further fine-tune on our NP annotations. We obtain the best scores with the latter. This demonstrates that first creating synthetic discourse annotations for NP and then proceeding with fine-tuning an English classifier model helps in classifying NP implicit discourse relations, even if the synthetically obtained annotations are over a different domain than that of the evaluation data.

We describe the methods used to generate a corpus of NP discourse relation annotations, thereby experimenting with different word alignment tools and methods to fine-tune these tools. The resulting corpora and the code to reproduce our results are published on GitHub\footnote{\url{https://anonymous.4open.science/r/NigerianPidginIDRC-331E}}. 

We hope that the procedures explained in this paper inspire others working on low-resource languages, specifically in the context of PLMs and LLMs, on tasks that are not easily improved by modern prompting procedures.

%% file: sections/limitations.tex

\section{Limitations}
\label{sec:limitations}

While we believe many procedures described are relevant for other low-resource languages, it's important to note that Nigerian Pidgin (NP) had English as its lexifier. Prior work \cite{lin2023low} has shown that English models perform better on NP than multilingual ones. In this paper, we find that fine-tuning an English classifier with NP data yields the best results, likely due to lexical similarities between English and NP. Results may differ for other low-resource languages.



Furthermore, due to the relatively large number of different system configurations and setups we used, we did not train the models multiple times. However, we averaged the evaluation results for all setups and found that the standard deviation was close to zero, indicating the stability of our main results

Our experiments were done on a Tesla V100-PCIE-32GB GPU. For other low-resource scenarios, with correspondingly limited resources (not just in terms of corpora, but also in terms of technical infrastructure available to the people working on those scenarios), reproducing our methods on, for example, a CPU, might not be feasible.

\section{Acknowledgments}

Muhammed Saeed is supported by the Konrad Zuse School of Excellence in Learning and Intelligent Systems (ELIZA) through the DAAD programme Konrad Zuse Schools of Excellence in Artificial Intelligence, sponsored by the Federal Ministry of Education and Research.

%% file: sections/appendix/appendix_a.tex
\section{Appendix A: NP PDTB Relation Sense Distribution}
\label{sec:appendix}

\begin{figure}[htp]
    \centering
    \begin{subfigure}{\linewidth}
        \centering
        \includegraphics[width=\linewidth]{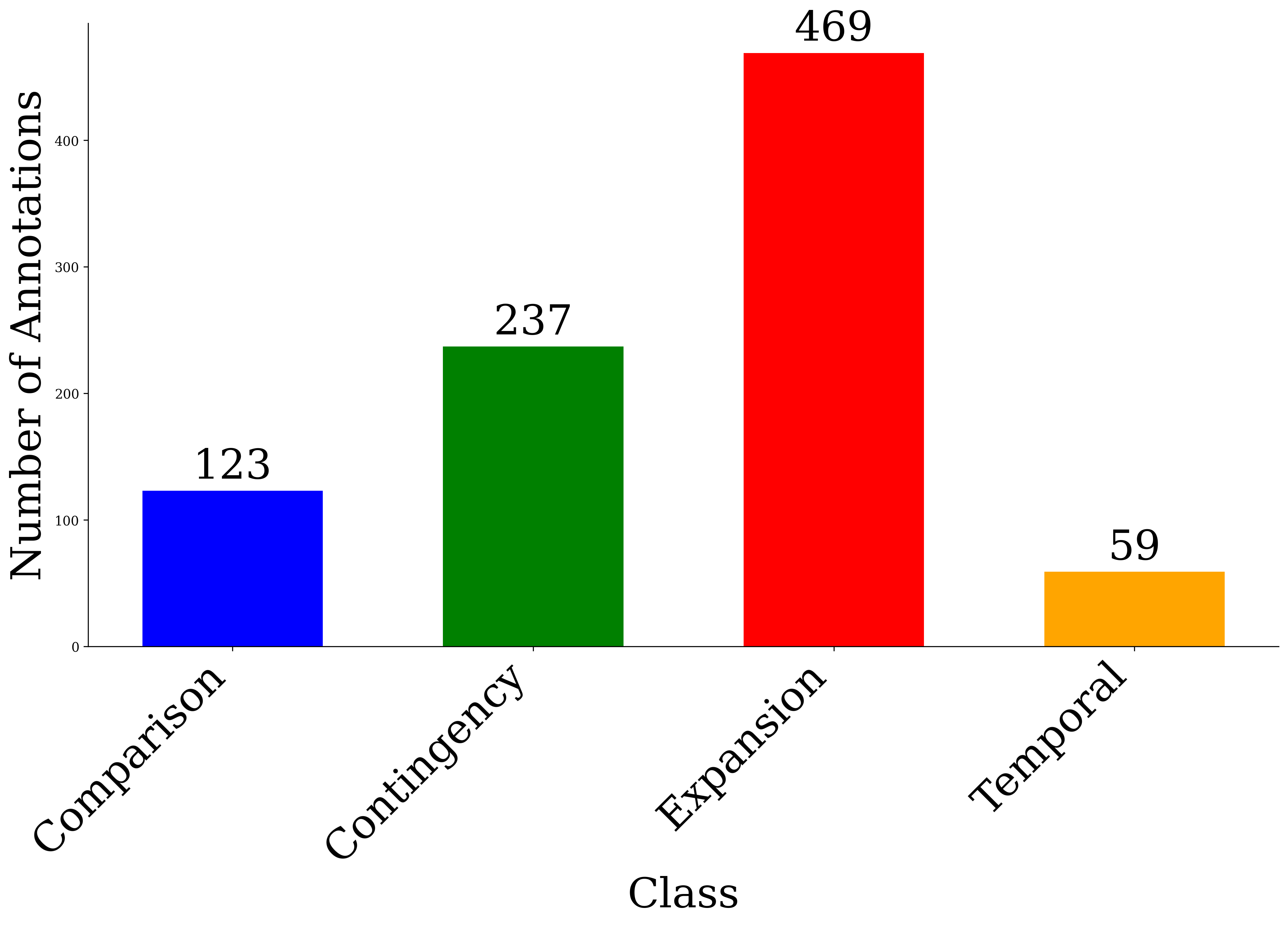}
        \label{fig:Awesomealignpft4way}
    \end{subfigure}
    \vspace{-1cm}  

    \begin{subfigure}{\linewidth}
        \centering
        \includegraphics[width=\linewidth]{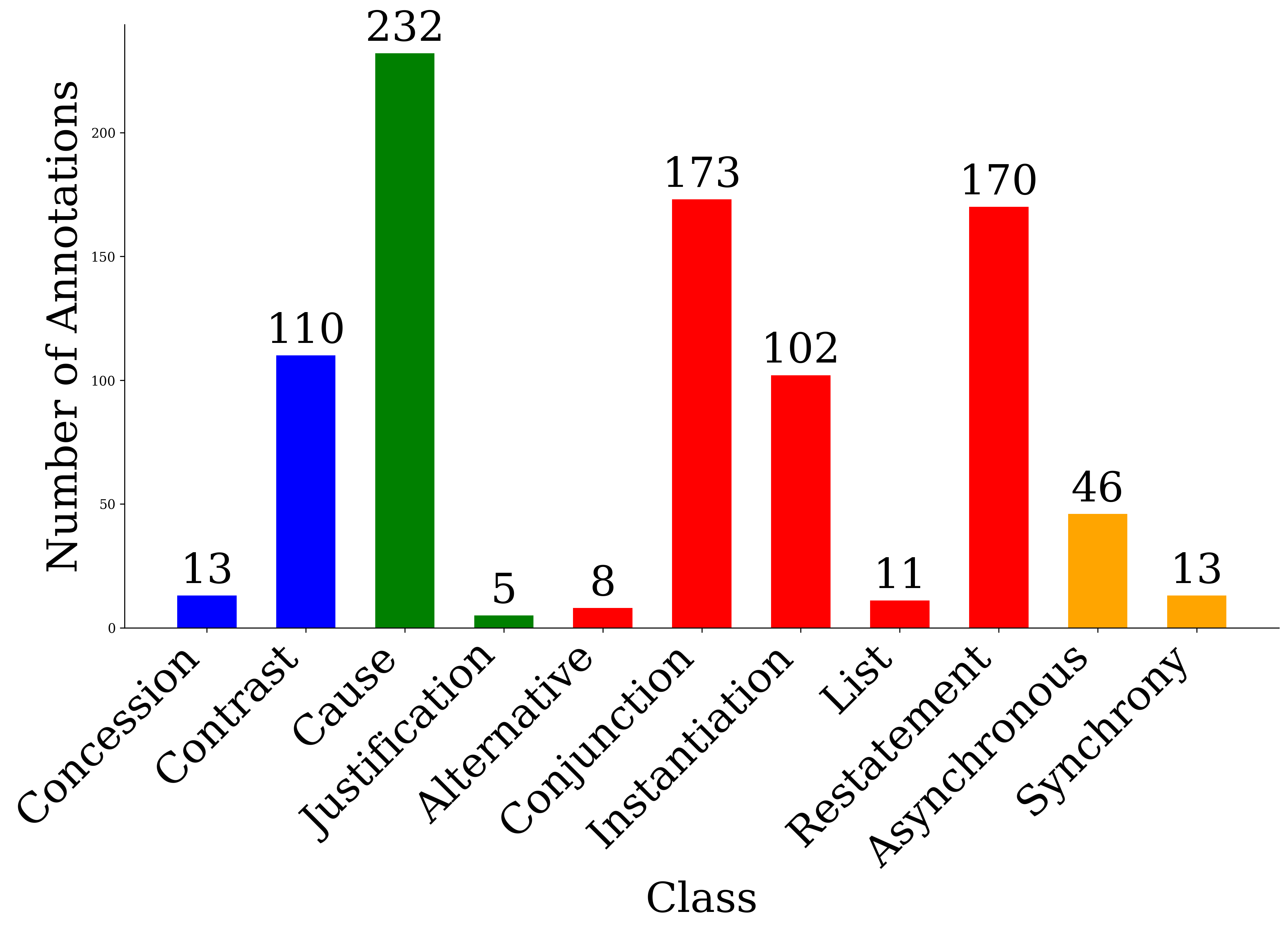}
        \label{fig:AwesomeAlignpft11way}
    \end{subfigure}
    \vspace{-0.5cm}  
    \caption{AWESoME+PFT NP PDTB relation sense distribution on top-level (top) and second-level (bottom).}
    \label{fig:awesomealign_pft_projected}
\end{figure}

Throughout our experiments, we have obtained the best results with the data set generated with the AWESoME word aligner. For 4-way classification, f$_{1}$ was best when deploying parallel fine-tuning (PFT), for 11-way classification, the data set created without PFT resulted in the best f$_{1}$ score. Although the number of relations (and therefore relation sense distribution) slightly differs for the different versions of our corpus (see Table \ref{tab: full transfer data-stats} in Section \ref{sec:data}), the sense distribution is very comparable for all versions of the corpus. Figure \ref{fig:awesomealign_pft_projected} illustrates the sense distribution for the corpus created with AWESoME+PFT, on the top level and second level of the PDTB hierarchy, for the data we used to train (from scratch) or fine-tune (CAFT) our Naija model. This, of course, closely resembles the sense distribution of \textit{implicit} relations in the original, English PDTB, but we include it to provide an idea of the distributions for readers not familiar with the PDTB.

%% file: sections/appendix/appendix_b.tex
\section{Appendix B: Machine Translation Engines}

For both the zero-shot set-up and the creation of our synthetic corpora, we rely on the Machine Translation system proposed by \citet{lin2023low}. We have experimented with other systems as well, notably PLM4MT \cite{tan2021msp} and Llama2 \cite{touvron2023llama}. For PLM4MT, we obtained a BLEU score of 20, compared to the 36 reported in \citet{lin2023low}. Since the approach based on Llama2 often generated additional (formatting-related) output characters, post-processing was necessary before a BLEU score could be calculated. Also, a frequent error faced during translation was the model's simply outputting the original (English) input text as the (incorrect) NP target text. Because we achieved better results with the model from \citet{lin2023low}, we refrained from doing post-processing and have no BLEU score for Llama2.

%% file: tables/ablation_table.tex
\begin{table*}[h!]
\centering

\begin{adjustbox}{width=1\textwidth}
\begin{tabular}{lcccc}
\hline
\textbf{Method} & \textbf{4-way f$_{1}$} & \textbf{4-way Accuracy} & \textbf{11-way f$_{1}$} & \textbf{11-way  Accuracy} \\
\hline
\textbf{\hspace{0.5cm}RELATION-BASED} \\ \hline
\textbf{\hspace{1cm}Naija model with CAFT, large T5 model} \\ \hdashline
\textbf{\hspace{1.5cm}AWESoME} & 0.433 & 0.585 & 0.325 & 0.444 \\
\textbf{\hspace{1.5cm}AWESoME+PFT} & 0.437 & 0.573 & \textbf{0.327} & 0.440 \\
\textbf{\hspace{1.5cm}SimAlign} & \textbf{0.461} & \textbf{0.631} & 0.246 & 0.429 \\
\textbf{\hspace{1.5cm}Simalign+CAT} & 0.411 & 0.571 & 0.236 & 0.425 \\
\textbf{\hspace{1.5cm}Giza-py} & 0.431& 0.566 & 0.289 &\textbf{0.446} \\
\hdashline
\textbf{\hspace{1cm}Naija model with CAFT, base T5 model} \\ \hdashline
\textbf{\hspace{1.5cm}AWESoME} & 0.352 & 0.513 & 0.197 & 0.360 \\
\textbf{\hspace{1.5cm}AWESoME+PFT} & 0.381 & 0.550 & 0.241 & 0.396 \\
\textbf{\hspace{1.5cm}SimAlign} & 0.382 & 0.614 & 0.228 & 0.379 \\
\textbf{\hspace{1.5cm}SimAlign+CAT} & 0.354 & 0.579 & 0.229 & 0.367 \\
\textbf{\hspace{1.5cm}Giza-py} & 0.388&0.581 &0.255 & 0.425\\
\hdashline
\textbf{\hspace{1cm}Naija model from scratch, large T5 model} \\ \hdashline
\textbf{Naija model from scratch} & 0.159 & 0.340 & 0.113 & 0.317 \\ 
\textbf{\hspace{1.5cm}AWESoME+PFT} & 0.364& 0.519 & 0.243 & 0.400 \\
\textbf{\hspace{1.5cm}SimAlign} & 0.296& 0.521 & 0.195 & 0.327 \\
\textbf{\hspace{1.5cm}SimAlign+CAT} & 0.324 & 0.445 & 0.167 & 0.335 \\
\textbf{\hspace{1.5cm}Giza-py} & 0.327& 0.389 & 0.216 &0.310 \\
\hdashline
\textbf{\hspace{1cm}Naija model from scratch, base T5 model} \\ \hdashline
\textbf{\hspace{1.5cm}AWESoME} & 0.354 & 0.515 & 0.278 & 0.392\\
\textbf{\hspace{1.5cm}AWESoME+PFT} & 0.357 & 0.538 & 0.246 & 0.340 \\
\textbf{\hspace{1.5cm}SimAlign} & 0.375 & 0.610 & 0.230 & 0.360 \\
\textbf{\hspace{1.5cm}SimAlign+CAT} & 0.229 & 0.377 & 0.183 & 0.321 \\
\textbf{\hspace{1.5cm}Giza-py} &0.335& 0.464 &0.256 &0.360\\

\hline
\hline
\textbf{\hspace{0.5cm}ALIGNMENT-BASED} \\ \hline
\textbf{\hspace{1cm}Naija model with CAFT, large T5 model} \\ \hdashline
\textbf{\hspace{1.5cm}AWESoME} & 0.408 & 0.575 & 0.304 & 0.416 \\
\textbf{\hspace{1.5cm}AWESoME+PFT} & 0.371 & 0.481 & 0.289 & 0.400 \\
\textbf{\hspace{1.5cm}SimAlign} & 0.362 & 0.498 & 0.259 & 0.396 \\
\textbf{\hspace{1.5cm}SimAlign+CAT} & 0.358 & 0.506 & 0.259 & 0.404 \\
\textbf{\hspace{1.5cm}Giza-py} & 0.368 & 0.552&0.245 &0.371 \\
\hdashline
\textbf{\hspace{1cm}Naija model with CAFT, base T5 model} \\ \hdashline
\textbf{\hspace{1.5cm}AWeSoME} & 0.373 & 0.479 & 0.289 & 0.356 \\
\textbf{\hspace{1.5cm}AWESoME+PFT} & 0.410 & 0.596 & 0.258 & 0.423 \\
\textbf{\hspace{1.5cm}SimAlign} & 0.356 & 0.496 & 0.258 & 0.383 \\
\textbf{\hspace{1.5cm}SimAlign+CAT} & 0.343 & 0.479 & 0.236 & 0.402 \\
\textbf{\hspace{1.5cm}Giza-py} &0.356 &0.577 & 0.190& 0.321\\
\hdashline

\textbf{\hspace{1cm}Naija model from scratch, large T5 model} \\ \hdashline
\textbf{\hspace{1.5cm}AWESoME} &0.228 & 0.427 & 0.144 & 0.329 \\
\textbf{\hspace{1.5cm}AWESoME+PFT} & 0.128  & 0.300 & 0.094 & 0.296 \\
\textbf{\hspace{1.5cm}SimAlign} & 0.267 &0.540 & 0.126 & 0.285 \\
\textbf{\hspace{1.5cm}SimAlign+CAT} &0.127 & 0.300 & 0.090& 0.294 \\
\textbf{\hspace{1.5cm}Giza-py} & 0.122 & 0.296 & 0.085 &0.292 \\
\hdashline
\textbf{\hspace{1cm}Naija model from scratch, base T5 model} \\ \hdashline
\textbf{\hspace{1.5cm}AWESoME} &0.307 & 0.396 & 0.235 & 0.285 \\
\textbf{\hspace{1.5cm}AWESoME+PFT} & 0.392  & 0.535 & 0.335 & 0.398 \\
\textbf{\hspace{1.5cm}SimAlign} & 0.304 & 0.375 & 0.204 & 0.315 \\
\textbf{\hspace{1.5cm}SimAlign+CAT} &0.205 & 0.390 &  0.126& 0.306 \\
\textbf{\hspace{1.5cm}Giza-py} & 0.138  & 0.315 & 0.101 &0.304 \\
\hline
\hline

\hline
\end{tabular}
\end{adjustbox}
\caption{Results for different versions of our NP training corpus, obtained by different word alignment set-ups. CAFT stands for continuous adaptive fine-tuning (applied to model used by DiscoPrompt), CAT stands for cross-lingual adaptive training (applied to model used by SimAlign), PFT stands for parallel fine-tuning (applied to model used by AWESoME).}

\label{tab:ablation}
\end{table*}